\documentclass{article}
\usepackage{spconf,amsmath,graphicx, multirow}
\usepackage{comment}
\usepackage{amsmath,graphicx}
\usepackage{graphicx}
\usepackage{caption}
\usepackage{subcaption}

\title{DEVELOPING REAL-TIME STREAMING TRANSFORMER TRANSDUCER FOR SPEECH RECOGNITION ON LARGE-SCALE DATASET}

\name{Xie Chen*$^1$, Yu Wu*$^2$, Zhenghao Wang$^1$, Shujie Liu$^2$, Jinyu Li$^1$\thanks{*Equal Contribution}}
\address{$^1$Microsoft Speech and Language Group \\
$^2$Microsoft Research Asia}
\begin{document}
\ninept
\maketitle
\begin{abstract}
Recently, Transformer based end-to-end models have achieved great success in many areas including speech recognition. However, compared to LSTM models, the heavy computational cost of the Transformer during inference is a key issue to prevent  their applications.   In this work, we explored the potential of Transformer Transducer (T-T) models for the fist pass decoding with low latency and fast speed on a large-scale dataset. We combine the idea of Transformer-XL and chunk-wise streaming processing to design a streamable Transformer Transducer model. We demonstrate that T-T outperforms the hybrid model, RNN Transducer (RNN-T), and streamable Transformer attention-based encoder-decoder model in the streaming scenario. Furthermore, the runtime cost and latency can be optimized with a relatively small look-ahead.
\end{abstract}
\section{Introduction}
\vspace{-0.1cm}
\label{sec:intro}
Great progress has been made to automatic speech recognition (ASR) with end-to-end (E2E) models \cite{chan2016listen, prabhavalkar2017comparison, battenberg2017exploring,  rao2017exploring, chiu2018state, Li18CTCnoOOV, he2019streaming, Li2020Developing, shi2020emformer}. Currently, Transducer (e.g., recurrent neural network Transducer (RNN-T) \cite{Graves-RNNSeqTransduction}) and Attention-based Encoder-Decoder (AED) \cite{chan2016listen, Attention-bahdanau2014, Attention-speech-chorowski2015} are two most popular types of E2E methods. AED models achieve very good performance thanks to the attention mechanism, however they are not streaming in nature by default 
and there are several studies towards that direction, such as monotonic chunk-wise attention \cite{DBLP:conf/iclr/ChiuR18} and triggered attention \cite{DBLP:conf/icassp/MoritzHR19, wang2020reducing}. In contrast, because of the streaming nature, transducer models especially RNN-T received a lot of attention for industrial applications and have also managed to replace traditional hybrid models for some cases \cite{he2019streaming, Li2020Developing, jain2019rnn}. 

While RNN with long short-term memory (LSTM) \cite{Hochreiter1997long} units was widely used in the E2E works, the transformer architecture with self attention \cite{vaswani2017attention} has recently become the fundamental building block for E2E models \cite{dong2018speech, karita2019comparative, Li2020Comparison}. In addition to accuracy, streaming recognizer and runtime computational cost are  two crucial factors for deploying high quality automatic speech recognition (ASR) system in industry. In this work, we focus on developing streaming Transformer transducer (T-T) \cite{zhang2020transformer}  and its variation with high accuracy and low computation cost. The computational cost of Transformer Transducer grows significantly with respect to the input sequence length, which obstacles the practical use of T-T. Recently conformer Transducer (C-T)  \cite{gulati2020conformer} was proposed to further improve T-T, but it is not streamable because its encoder has attention on full sequence. 

Existing methods may partially address these issues, but have their own drawbacks. 1) Time-restricted method \cite{zhang2020transformer,DBLP:journals/corr/abs-2001-02674,yu2020universal,tripathi2020transformer} simply masks left and right context in Transformer to control time cost. As the reception field grows linearly with the number of transformer layers, a large latency is introduced with the strategy. 2) chunk-wise method \cite{tian2020synchronous,wang2020reducing} segments the input into small chunks and operates  speech recognition on each chunk. However, the accuracy drops as the relationship between different chunks are ignored. 3) Memory based method \cite{wu2020streaming,inaguma2020enhancing} employs a contextual vector to encode history information while reducing runtime cost by combining with chunk-wise method. However, the method breaks the parallel nature of Transformer in training, requiring a longer training time.

In this paper, our goal is to develop streaming Transformer and Conformer Transducer models that can be operated in real time. We wish to reach a balance between training cost, runtime cost, and accuracy. We combine Transformer-XL and chunk-wise processing to handle streaming scenario, and there is no overlap between chunks in training to guarantee the training efficiency. We can finish training 65 thousand hours of anonymized training data in 2 days with mixed precision on 32 V100 GPUs. T-T and C-T outperform hybrid model, RNN-T, and streaming Transformer attention-based encoder-decoder model over 10$\%$ relative word error rate (WERR) on the streaming speech recognition evaluation. In term of runtime cost, our proposed method uses limited history while maintaining the same performance (1$\%$ WERR degradation). If a small look-ahead is allowed, such as 360ms, T-T reaches 0.25 real-time factor rate on CPU that satisfies the industry requirement of a real application.

\section{Model Structure}
\vspace{-0.2cm}

\subsection{Transducer Architecture}
In this paper, we investigate the use of transducer model \cite{graves2012sequence} for real-time and streaming speech recognition. 
A transducer has three components, an acoustic encoder network (encoder), a label predictor network (predictor), and a joint network.
The acoustic feature sequence $\textbf{x}_1^t$ is fed into the encoder to get encoder output 
$\textbf{f}_t$, correspondingly the previous label sequence $\textbf{y}_1^{u-1}$ are sent to the 
predictor to compute the predictor output $\textbf{g}_{u-1}$. The outputs of encoder and predictor are then
added by the joint network. A non-linear function, such as relu function, is applied before sending to softmax function to compute the probability distribution over the sentence piece vocabulary. 
The computation formulas in transducer could be written as below,
\begin{eqnarray}
    \textbf{f}_{t} &=& \text{encoder}(\textbf{x}_1^{t}) \nonumber \\
    \textbf{g}_{u-1} &=& \text{predictor}(\textbf{y}_1^{u-1}) \nonumber \\
    \textbf{h}_{t, u-1} &=& \text{relu}(\textbf{f}_t+\textbf{g}_{u-1})    \nonumber \\
    P(y_u|\textbf{x}_1^t, \textbf{y}_1^{u-1}) &=& \text{softmax}(W_o * \textbf{h}_{t, u-1}) 
    \label{eqn:1}
\end{eqnarray}

In practice, we can employ different architectures for encoder and predictor. For example, in \cite{graves2012sequence}, LSTM is used for both encoder and predictor and is widely known as RNN-T model structure. In this paper, we use transformer as the encoder, and LSTM as the predictor due to the consideration of speed and memory cost.
\subsection{Transformer and Conformer}
\label{sec:transformer}
In the past several years, the transformer model \cite{vaswani2017attention} has proven to present significant performance improvement over LSTM in
a range of tasks \cite{dai2019transformer, wang2020transformer}. Very recently, the transducer using transformer were proposed and reported to outperform LSTM based transducer models 
\cite{wang2020transformer, zhang2020transformer, yeh2019transformer}. In literature, to distinguish RNN-T which adopted RNN as encoder and decoder, the transformer based transducer is normally called transformer transducer (T-T).

The transformer model adopts the attention
mechanism to capture the sequence 
information. Self-attention is used to 
compute the attention distribution over the input sequences with a dot-product similarity function,
which could be written as,
\begin{eqnarray} \small
    \alpha_{t, \tau} &=& \frac{\exp(\beta (W_q\textbf{x}_t)^T(W_k\textbf{x}_\tau))}{\sum_{\tau'}\exp(\beta (W_q\textbf{x}_t)^T (W_k\textbf{x}_{\tau'}))}  \nonumber \\ 
    &=& \text{Softmax} (\beta \textbf{q}_t^T \textbf{k}_\tau)  \nonumber \\ 
        \textbf{z}_t &=& \sum_{\tau} \alpha_{t\tau} W_v \textbf{x}_\tau \nonumber\\
        &=& \sum_{\tau} \alpha_{t\tau} \textbf{v}_\tau 
\label{eqn:2}
\vspace{-0.2cm}
\end{eqnarray}
where $\beta=\frac{1}{\sqrt{d}}$ is a scaling factor. The input vector $\textbf{x}_t$ is
sent to three different matrices and the outputs are used as query $\textbf{q}_t$, key $\textbf{k}_t$ and value $\textbf{v}_t$ respectively
in the attention module. In the transformer model, multi-head attention (MHA) is applied to further improve
the sequence model capacity, where multiple 
parallel self-attentions are applied on the input sequence and the outputs of each attention module are then concatenated. The range of
input sequence for the softmax function in Equation \ref{eqn:2} can be controlled by applying a mask. If we only want to use the current and previous frames $\textbf{x}_1^t$ to compute $\textbf{z}_t$, the attention weights $\alpha_{t, \tau}$, where $\tau > t$, could be masked to be 0.  Hence, the use of mask provides a flexible approach to decide the scope of input sequence for computation. The details of the mask design used in this paper can be found in Section \ref{sec:mask}. In each transformer layer, it also contains two fully-connected feed-forward networks (FFN), a nonlinear activation, layer normalization, and residual connections. A transformer-based audio encoder usually stacks multiple transformer layers, e.g. 18 layers.

In transformer model, the position embedding is normally used to explicitly model the ordering information of input sequence. Relative
position embedding was found to yield better performance compared to absolute position embedding \cite{shaw2018self, dai2019transformer, wang2020transformer}. The motivation is the offset between two frames should be considered in the attention weight calculation, and the offset is modeled by the relative position embedding. For efficiency, we use a simple but effective relative position embedding, which is formulated as  

\begin{equation} \small
    \textbf{z}_t = \text{Softmax}(\beta \mathbf{\textbf{q}_t^T}(\mathbf{\textbf{k}_\tau+\textbf{p}_{t, \tau}}) ) \mathbf{\textbf{v}_\tau \label{self_att}} 
\end{equation} 
where $\textbf{p}_{t, \tau}$ is the relative position embedding obtained from a lookup table.
The implementation is more efficient and memory friendly than the relative position embedding used in \cite{dai2019transformer}. 

The Transformer model captures global context, but the local information does not model very well. A few recent work \cite{gulati2020conformer} shows that the marriage of CNN and Transformer improves the SR performance. Among them, Convolution augmented Transformer (a.k.a Conformer) \cite{gulati2020conformer} is a typical one, which inserts a special CNN based structure into each Transformer block, which achieves state-of-the-art performance on Librispeech.  We adopt the original structure of \cite{gulati2020conformer} in our implementation but changing the depth-wise CNN to causal depth-wise CNN to avoid extra latency.

\section{Streaming Transformer-Transducer in Real-Time}
\label{sec:approach}
\subsection{Challenges in Streaming Scenario}

During inference, the computational cost is a potential issue if the full history is used for the computation of each frame. The computation of $\textbf{z}_t$ in Equation \ref{eqn:2} will increase linearly with  $t$ as it needs to compute the attention weights from the first to the current frames. As a result, it makes the overall computation complexity quadratic, which is not affordable for long utterances. 
Furthermore, there is a trade-off between the model accuracy and latency, if we could allow several look-ahead frames for the computation of the current frame, performance improvement could be achieved. The latency controlled technique has been applied widely in Hybrid systems \cite{zhang2016highway} and also introduced into transducer models recently \cite{yeh2019transformer, zhang2020transformer}.
In order to make computation feasible and achieve better performance, \textbf{truncated history} and \textbf{limited future information} can be used for transformer transducer.
In \cite{yeh2019transformer, zhang2020transformer}, the authors set the attention mask to allow a
specific number of contextual frames in each transformer layer for both history and the future. This is able to reduce the computation in each layer efficiently. However, there is a potential drawback for the fixed number of future context. With the increase of transformer layers, the number of future context increases linearly. If the transformer has 18 layers and the future context is 5 frames per layer, a latency of 90 frames will be introduced as a result.

\subsection{Attention Mask Design for Training Streaming Models }
\label{sec:mask}

\begin{figure}
     \centering
     \begin{subfigure}[b]{0.45\textwidth}
         \centering
         \includegraphics[width=7.0cm]{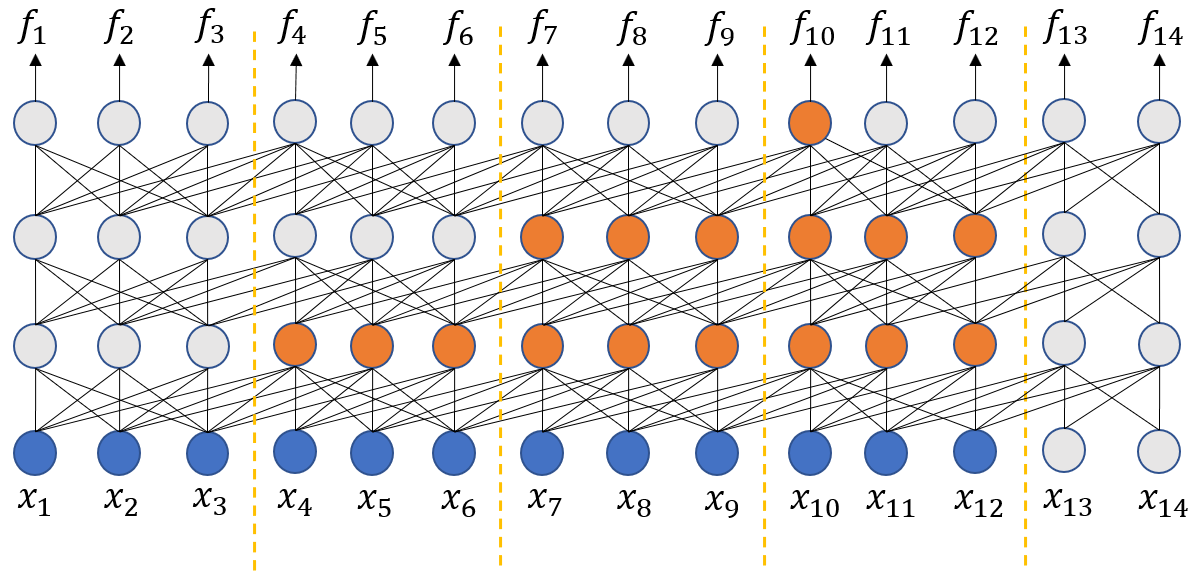}
         \caption{The reception field of  position $x_{10}$. The left reception field grows with the number of Transformer layers, whereas the right reception field does not.}
         \label{fig:attn_mask_1}
     \end{subfigure}
     \hfill
     \begin{subfigure}[b]{0.45\textwidth}
         \centering
         \includegraphics[width=7.0cm]{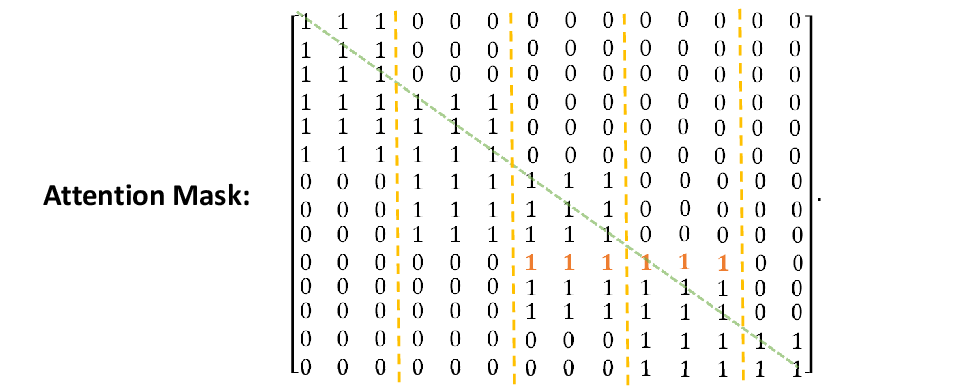}
         \caption{Attention mask matrix $M$ per layer. If $M(i,j)$ is 1, then the $j$th input will be used for computation in $i$th frame. }
         \label{fig:attn_mask_2}
     \end{subfigure}
     \hfill
     \caption{ An example of our mask strategy. Chunk size and history window size are both 3 in the example. }
        \label{fig:attn_mas}
        \vspace{-0.5cm}
\end{figure}

We design a simple but effective mask strategy to truncate history and allow limited future information. 
Transformer uses attention mechanism for sequence modeling and the attention mask can be applied on the attention weight matrix $\{\alpha_{t, \tau}\}$, to determine the range of input sequence involved for computation. When $\alpha_t,\tau$ is set to 0 by the attention mask, then the input at time $\tau$, $\textbf{x}_\tau$, won't be used for the computation of output $\textbf{z}_t$ at time $t$. The mask matrix is created with the following rules. First, the input acoustic feature sequence is segmented into chunks with
a specified chunk size.\footnote{ Chunks do not have overlap for efficient training.} Then we construct the matrix according to the following rules: 1) Frames in the same chunk can see each other. For example, when computing the encoder output for  frame $\textbf{x}_{10}$, all frames belonging to the same chunk, including $\textbf{x}_{10}, \textbf{x}_{11}, \textbf{x}_{12}$, are considered in the attention calculation. $x_{10}$ can see two future frames, while $x_{12}$ has zero frame lookahead. The average number of lookahead frames for each frame is the half of the chunk size.  2) If two frames are in different chunks, the left one cannot see the right one in attention calculation, even if the offset is one. By this means, the chunk boundary will strictly restrict reception field, avoiding the linear reception field growing as the model becomes deeper. 3) If two frames are in different chunks, the right one can see the left one if their distance is less than the history window size. This method makes the left reception field grow linearly (history window size) with the model becoming deeper. Figure \ref{fig:attn_mask_2} gives an example of the streaming mask matrix.  This attention mask is shared across all transformer layers.

Figure \ref{fig:attn_mask_1}  gives the reception field of the position $x_{10}$. We can find that the advantage of the masking strategy is that it allows the left context linearly increase while forbids the right reception field grows, so the model can use long history information while restrict future look-ahead. In the example, the left reception field grows 3 frames per layer, while the right context is restricted to $x_{12}$ for all layers. Moreover, the masking strategy is very flexible, it can simulate most of the possible scenarios we will use in practice. When chunk size is one and history window is infinity, it simulates the naive zero look-ahead scenario. 

\subsection{Inference Optimizations}

We use the following engineering optimizations for inference. 

\textbf{Caching:} given the input acoustic frame at time 
$t$, $\textbf{x}_t$, in order to compute the transformer encoder output $\textbf{f}_t$,
in each transformer layer, in addition to the linear matrix multiplications and 
non-linear functions, we also need to compute the 
attention weight over the input sequence and then 
sum over the weighted value vectors according to Equation 
\ref{eqn:2}.  In order to avoid repeated computation, some 
intermediate variables are cached. Specifically, in Equation 
\ref{eqn:2}, the key $\textbf{k}_{\tau}=W_k\textbf{x}_\tau$ and value 
$\textbf{v}_{\tau}=W_v\textbf{x}_\tau$ in each layer are cached once 
computed. As a result, we only need to calculate the query $\textbf{q}_{t}=W_q\textbf{x}_t$, key $\textbf{k}_{t}$ and value $\textbf{v}_{t}$, and then applied softmax function over the input sequences using the cached key and value, for the acoustic frame at  time $t$ in each layer. 
In addition, as we used truncated history for transformer in each frame,
the memory consumption for caching won't increase with the increase of $t$.
 
 \textbf{Chunk-wise compute:} if the latency of several frames are allowed, as shown in Figure \ref{fig:attn_mas}, we could group the input frames $[\textbf{x}_{10}, \textbf{x}_{11}, \textbf{x}_{12}]$ as a small minibatch and feed to the transformer encoder to compute $\textbf{f}_{10}, \textbf{f}_{11}$ and $\textbf{f}_{12}$ simultaneously. The key and value in each layer will be cached once computed, which will be used in the computation of the future frames. In this way, the efficient matrix operation could be applied, instead of applying the matrix-vector operation multiple times. It is worth mentioning that, for the transformer encoder with zero lookahead, we can manually introduce a latency of several frames and significant speedup could be achieved by applying this chunk-wise computation.
\vspace{-0.3cm}

\section{experiment}
\vspace{-0.2cm}
\label{sec:experiment}
\subsection{Experiment Setup}
We used 65 thousand (K) hours of transcribed 
Microsoft data as the training data. 
The test set covers 13 different application scenarios 
such as Cortana, far-field speech and call center, consisting of a total of 1.8 million (M) words. The word error rate (WER) averaged over all
test scenarios are reported. All the training and test data are anonymized
data with personally identifiable information removed. 4000 sentence pieces 
trained on
the training data transcription was used as vocabulary.
We applied a context window of 8 for the input frames to form a 640-dim feature as the input of transducer encoder and the frame shift is set to 30ms. Utterances longer than 30 second
were discarded from the training data.

For the RNN-T, the encoder contains 6 LSTM layers and the predictor consists of one embedding layer and 2 LSTM layers. The dimension of embedding and LSTM layers are set to 1024. In terms of T-T, 18 transformer layers with 720 hidden nodes and 1024 feedforward nodes were used as encoder; Same as RNN-T, 2 LSTM layers with 720 hidden nodes were used as predictor. C-T chose 640 as the hidden layer size to get similar model size as RNN-T and T-T. The kernel size in C-T is 3. Relative position encoding is used for T-T. All the models are trained from scratch and with mixed precision for efficient training. The number of model parameter for various transducer models is around 80M.

The runtime speed is evaluated on a single CPU machine containing 16 cores,
with Intel Xeon CPU e5-2620, 2.10 GHz, and 64GB memory. 
We randomly sample 500 utterances from the test set to measure the run-time factors. The average length of 12.7s for these utterances. 
Float precision is used for evaluating LSTM and transformer transducer models
in the following experiment without an explicit statement.
For the beam search of transducer decoder, $n$-best is set to 5 for all experiments.
In order to conduct efficient decoding for transducer models, an efficient transducer decoder based on beam search was implemented with C++. The details on the decoding algorithm can be found in  \cite{graves2012sequence}. The transducer models used in this paper, including LSTM, transformer based transducers, are trained with Pytorch, then exported using Just-In-Time (JIT) compilation with Libtorch. These JIT  exported models can be evaluated in the decoder implemented with C++ on CPU conveniently and efficiently. Real time factor (RTF) is used to evaluate efficiency.

\subsection{Evaluation Result}

\subsubsection{Evaluation with the zero look-ahead setting}
The zero look-ahead model is important for a real system, since many applications require the system give a quick response to a user's query. Thus, the first experiment compares the performance of RNN-T and T-T with zero look-ahead on the 1.8M testset. All models do not see any frame in the future and decode frame by frame. Table \ref{tab:1} presents the accuracy and runtime cost. We can observe that T-T and C-T significantly outperform RNN-T in terms of accuracy, and C-T is slightly better than T-T, which is consistent with previous literature. However, the RTF of T-T and C-T is much higher than 1 if full context is attended. When we truncate left history using the method proposed in section \ref{sec:approach}, we find that the model keeps almost the same performance, while reduces RTF significantly. It is also worth noting the truncated history reduced the memory consumption notably. The full context requires caching all keys and values in each frame and the memory grows linearly with the increase of audio frames, while truncated history keeps a fixed length of key and value vectors, such as 60 in this experiment.
It demonstrates that the truncated history could reduce the runtime cost and memory effectively without affecting WER performance. 




\begin{table}[t]
    \centering
    \begin{tabular}{c|c|c|c|c}
    \hline
           & \#hist   &  WER & \multicolumn{2}{|c}{RTF (\#thread)}   \\

                             &   \#frames  &  (\%)     & 1 & 4 \\
    \hline
         RNN-T         & $+\infty$ & 9.86 & 1.56 & 0.46\\
    \hline
          T-T  & $+\infty$  & 8.79 & 3.44  & 2.57 \\
    \hline
          T-T   & 60    & 8.88 & 2.38  & 1.75 \\
        \hline
          C-T  & $+\infty$  & 8.78 & 4.02  & 2.56 \\
    \hline
          C-T   & 60    & 8.80 & 2.41  & 1.83 \\
    \hline
    \end{tabular}
    \caption{Model comparison for the zero look-ahead setting.}
    \label{tab:1}
    \vspace{-0.6cm}
\end{table}


However,  the RTF of T-T is still higher than 1 even if the model uses limited context in the zero look-ahead scenario. We find that the bottleneck of T-T is the encoder runtime cost, which occupies about 90$\%$ time in the whole inference stage. The frame-by-frame computation is time-consuming as it does not fully utilize the parallel computation of transformer models. Motivated by this, we introduce a trade-off between the latency and runtime cost by grouping several frames to form a batch for computation.
For RNN-T, we could also form the batch and feed it to the LSTM based audio encoder. Due to the recurrent connection in LSTM, it can only partially parallelize the computation and the speedup is expected to be slower than Transformer.
 Table  \ref{tab:2} reports RTF across different batch sizes. With a larger batch size, a faster decoding speed can be achieved. If we encode 2 frames (60ms latency) for each computation, RTF is less than 1. The RTF can be further reduced to 0.2 for T-T when the batch size is 15.
 However, 15 batch size introduces a latency of 450ms (15*30ms). 
 Moreover, Table \ref{tab:2} indicates that the transformer is more suitable than RNN-T for batch operation, and an appropriate batch size makes T-T achieve similar RTF with RNN-T.

 \begin{table}[htbp]
     \centering
     \begin{tabular}{c|c|c|c|c|c|c|c}
     \hline
     & \#hist      &  WER  & \multicolumn{5}{|c}{RTF (\#batch size)} \\
     \cline{4-8}
               &  len         & (\%) & 1 & 2   & 5  & 10 & 15 \\
     \hline
     RNN-T         &  $+\infty$ & 9.86& 0.46 & 0.31  & 0.26 & 0.21 & 0.20\\
     \hline
      T-T & 60 &  8.88  & 1.75 & 0.69  & 0.38 & 0.26 & 0.19 \\
           \hline
      C-T & 60 &  8.80 & 1.83 & 0.95  & 0.48  & 0.36 & 0.25  \\
     \hline
     \end{tabular}
    \caption{WER and RTF results of accumulating various number of frames for batch computation for RNN-T, T-T and C-T trained with zero lookahead.  4 threads are used for evaluation.}
     \label{tab:2}
     \vspace{-0.5cm}
\end{table}

\subsubsection{Evaluation with the small look-ahead setting}
 
 According to the zero look-ahead experiment, T-T has to trade latency for less computation cost. An absolute zero look-ahead is impossible for T-T, and it has to encode frames with a batch. Therefore, an idea is to do speech recognition (SR) with a small look-ahead, which enables the chunk-wise decoding in a natural way. When T-T and C-T take 24 frames lookahead, it generates a latency window between [0, 24] frames, and the averaged latency turns to be 24*30 ms /2=360ms, while RNN-T did not apply the chunk-based decoding and its latency is fixed as 12 frames (360ms). We also copy the number from \cite{Li2020Comparison} to show the performance hybrid system, streamable Transformer Seq2Seq, and offline Transformer Seq2Seq model on the test set, where the hybrid model is a highly optimized contextual layer trajectory LSTM (cltLSTM) \cite{li2019improving}, the streamable Transformer Seq2Seq is based on the chunk-wise trigger attention method, \cite{DBLP:conf/icassp/MoritzHR19, wang2020reducing}, and the streamable RNN S2S is based on MoCha\cite{DBLP:conf/iclr/ChiuR18}. 
 
 Table \ref{tab:3} presents the performance of different models with a small lookahead.
  The results indicate that transducer models are more powerful than S2S in the streaming scenario, since RNN-T and T-T outperform streamable RNN S2S and streamable Transformer S2S respectively. T-T and C-T are the better choices for the scenario with a small look-ahead, as it shows strong accuracy while achieves an acceptable runtime cost. One unanticipated finding was that T-T with a small lookahead is very close to the performance of T-T using the entire utterance, which suggests that our simple strategy for T-T can avoid huge performance drop compared to the offline model.

\begin{table}[t] 
    \centering
    \begin{tabular}{c|c|c|c|c|c}
    \hline
            & \#hist  & \#lookahead  &  WER & \multicolumn{2}{|c}{RTF (\#thread)}   \\
    \cline{5-6}
                       &  frame          &   (ms)  &     (\%) & 1 & 4 \\
    \hline
         Hybrid        & $+\infty$ & 480 & 9.34  & -  & -  \\
     \hline
         RNN S2S         & $+\infty$ & 720 & 9.61  & -  & -  \\
    \hline
         Trans. S2S         & $+\infty$ & [480, 960] & 9.16  & -  & -  \\
    \hline
         Trans. S2S         & $+\infty$ & $+\infty$ & 7.82  & -  & -  \\
    \hline  \hline
         RNN-T         & $+\infty$ & 360 & 9.11 & 1.52 & 0.43\\
        \hline
         T-T   & 60&   [0,720]    & 8.28 &   0.40 & 0.16  \\
    \hline
        C-T   & 60&   [0,720]    & 8.19 &  0.45 & 0.22  \\
    \hline
         T-T   & $+\infty$ &   $+\infty$   & 7.78 & 0.39  & 0.15 \\
    \hline
        C-T   & $+\infty$ &   $+\infty$   & 7.69 & 0.36  & 0.15 \\
    \hline
    \end{tabular}
    \caption{WER and RTF comparison for different streaming models with lookahead. The first block results using Hybrid and S2S models are from \cite{Li2020Comparison}.}
    \label{tab:3}
\vspace{-0.5cm}
\end{table}

\subsubsection{8-bit Optimization}
The final experiment compared the effect of INT8 quantization. The INT8 quantization can reduce memory consumption and speed up inference effectively while keeping the performance. Table \ref{tab:4} shows the WER and speed results by using INT8 with 1 thread. INT8 presents 3.6 times speedup for RNN-T without affecting WER performance. In contrast, INT8 introduces slight WER degradation on T-T and C-T and
yields about 2 times speedup. One possible explanation is that the softmax in Transformer layer is still operated in float precision, while softmax is computationally expensive on CPU.

\begin{table}[htbp]
    \centering
    \begin{tabular}{c|c|c|c}
    \hline
     &  Precision & WER (\%)  &  RTF \\
    \hline
    RNN-T   & float32     &  9.11   & 1.56   \\
           & int8      &  9.13     & 0.43  \\
    \hline
    T-T   & float32  & 8.28  &  0.40\\
                  &  int8  &  8.50  &  0.22      \\
    \hline
    C-T     & float32  & 8.19    & 0.45 \\
                  & int8   &  8.40    & 0.26  \\
    \hline
    \end{tabular}
    \caption{WER and speed results with INT8 quantization for Transducer models with lookahead.}
    \label{tab:4}
    \vspace{-0.5cm}
\end{table}
\vspace{-0.2cm}

\section{conclusion}
\vspace{-0.1cm}
\label{sec:conclusion}
We develop streaming T-T and C-T speech recognition model for real-time speech recognition, in the hope that the powerful Transformer encoder and streaming natural transducer architecture could take advantages from each other. We combine the idea of Transformer-XL and chunk-wise streaming processing to avoid latency grows linearly with the number of transformer layers. The experiment results show that T-T and C-T outperform hybrid model, RNN-T model, and streamable Transformer AED model in terms of accuracy in the streaming scenario. T-T and C-T can achieve comparable or better RTF compared to RNN-T given a small latency.

\vfill\pagebreak

\bibliographystyle{IEEEbib}
\bibliography{refs}

\end{document}